\documentclass[letterpaper, 10 pt, conference]{ieeeconf}  %

\IEEEoverridecommandlockouts                              %

\usepackage{cite}
\usepackage{amsmath,amssymb,amsfonts}
\usepackage{algorithmic}
\usepackage{graphicx}
\usepackage{textcomp}
\usepackage[table]{xcolor}
\usepackage{xurl}
\usepackage{multirow}
\usepackage{pifont}
\usepackage{xparse}

\def\BibTeX{{\rm B\kern-.05em{\sc i\kern-.025em b}\kern-.08em
    T\kern-.1667em\lower.7ex\hbox{E}\kern-.125emX}}

\title{ \bf DESCENT: Directed Edge Scene Encoding for Airport Surface Movement Prediction}

\author{Alexander Prutsch$^{1}$, David Schinagl$^{1}$ and Horst Possegger$^{1}$%
\thanks{$^{1}\,$Corresponding author: {\tt alexander.prutsch@tugraz.at}. All authors are with the Institute of Visual Computing, Graz University of Technology, Austria. 
This work was partially funded by the Austrian Research Promotion Agency (FFG) under the Take Off project SAFER (894164).}
}

\begin{document}

\def\mn{DESCENT}
\newcommand{\cmark}{\ding{51}} %
\newcommand{\xmark}{\ding{55}} %
\newcommand{\aptodo}[1]{\begingroup\color{black}#1\endgroup}
\newcommand{\apnew}[1]{\begingroup\color{black}#1\endgroup}
\definecolor{rcol}{rgb}{0.9,0.9,0.9}
\newcommand{\sv}[1]{\underline{#1}}
\newcommand{\bv}[1]{\textbf{#1}}
\newcommand{\ourrow}[0]{\rowcolor{rcol}\mn~(Ours)}

\def\eg{\emph{e.g.,}~} \def\Eg{\emph{E.g.,}~}
\def\ie{\emph{i.e.,}~} \def\Ie{\emph{I.e.,}}
\def\fe{\emph{f.e.,}~} \def\Fe{\emph{F.e.,}}
\def\cf{\emph{cf.}~} \def\Cf{\emph{Cf.}}
\def\etc{\emph{etc}\onedot} 
\def\vs{\emph{vs}\onedot}
\def\wrt{w.r.t\onedot~} 
\def\dof{d.o.f\onedot}
\def\etal{\emph{et al}.~}
\def\etc{\emph{etc}\dots~}

\maketitle
\thispagestyle{empty}
\pagestyle{empty}

\begin{abstract}
Advanced automation is a key technology for enhancing the safety of ground operations amidst the increasing density of commercial air traffic. While motion forecasting is a well-studied task in autonomous driving, its application to airport surface movements remains underexplored.
To enable efficient and accurate prediction in this domain, we propose \mn, a transformer-based architecture designed to handle heterogeneous dynamics and strict topological constraints.
Our approach features a Potential Reachable Set (PRS) context sampling mechanism that adaptively collects airfield environment context across diverse operational phases.
Combined with a detection transformer-based decoder, \mn~generates accurate trajectory forecasts.
Extensive evaluations on the Amelia-10 benchmark demonstrate significant performance improvements over state-of-the-art baselines.
These gains are especially pronounced in safety-critical scenarios, where our domain-aware sampling provides critical long-horizon context necessary for safe navigation.

\end{abstract}

\section{INTRODUCTION}

The continuous growth of commercial air traffic has led to a substantial increase in airport surface movements~\cite{icao2025state}.
This rise has resulted in higher runway occupancy rates and increased complexity in traffic management.
Consequently, the frequency of critical situations such as runway incursions~\cite{jtsb2024aircraft, eurocontrol2024global} has increased, posing significant safety concerns~\cite{icao2025state, ember2023airline}.
These risks are further intensified by air traffic control staffing shortages~\cite{krolik2025air}.
Trajectory prediction offers a promising technological solution to mitigate these risks: by forecasting the future path of aircraft and ground vehicles, potential conflicts can be detected early, allowing the system to issue timely warnings to air traffic controllers or pilots.
For instance, motion forecasting can be used to verify runway clearance during an aircraft’s final approach or to detect unsafe situations caused by unexpected interventions, \eg an aircraft taxiing across an active runway.

Deep learning-based trajectory prediction has been extensively studied across multiple domains, \eg autonomous driving~\cite{shi2022motion, zhou2023query, song2024realmotion}, pedestrian motion forecasting~\cite{salzmann2020trajectron, shi2023trajectory, bae2024singulartrajectory, fu2025moflow}, and airborne aircraft trajectory prediction~\cite{patrikar2022trajair, yin2023context, yin2025aircraft}.
In particular, much of the recent progress has been driven by advances in autonomous driving, where trajectory prediction is a standard component of the vehicle control stack.
By anticipating the future motions of surrounding traffic participants, \eg other vehicles, prediction modules support safe and efficient ego-motion planning.
Common architectures encode historical agent motions together with map context, before learning interactions among scene elements and generating multimodal trajectory hypotheses.
In principle, this paradigm can also be applied to airport surface operations, where runways and taxiways provide structured map context and the System Wide Information Management~(SWIM)~\cite{meserole2006system} provides motion data.
Transferring these predictive capabilities to the aviation domain can help to increase airport safety through semi-automated airfield operations and advanced robotic assistance for pilots.
Amelia-TF~\cite{navarro2024amelia} demonstrates the feasibility of autonomous driving-inspired prediction architectures for airport surface operations.
It is introduced as a baseline within the Amelia framework~\cite{navarro2024amelia}, which provides a large-scale airport surface movement dataset collected at 42 airports (Amelia-42), as well as a benchmark subset (Amelia-10).
Together, they enable the development of data-driven trajectory prediction methods for commercial aviation.

\begin{figure}[t]
    \centering
    \includegraphics[trim={3cm, 2.85cm, 0cm, 1.4cm}, clip, width=0.9\linewidth]{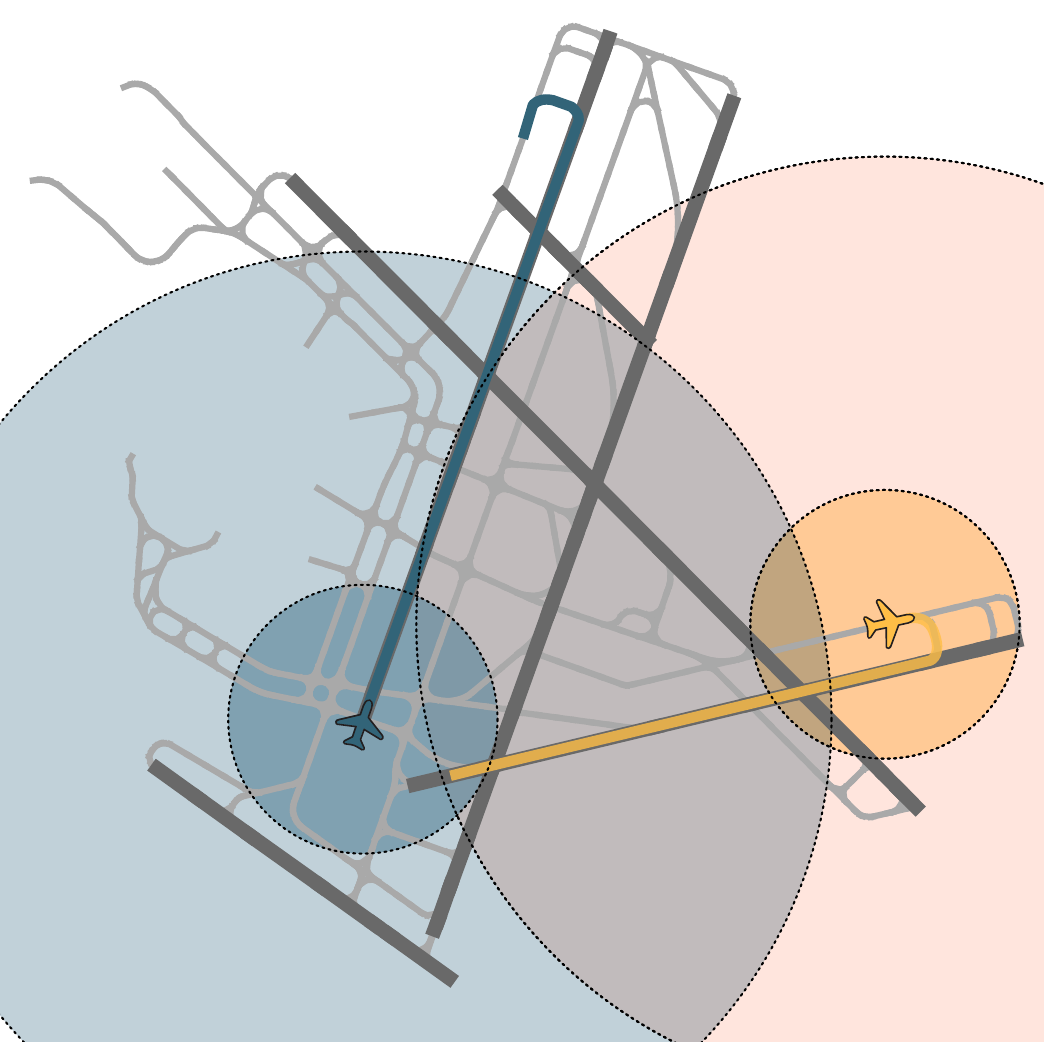}
    \vspace{-0.15cm}
    \caption{
    We illustrate two aircraft at Logan International Airport (KBOS) with future movements indicated by colored paths.
    Runways are shown in dark gray and taxiways in light gray.
    Standard map-sampling strategies -- such as top-$k$ nearest elements or fixed circular regions of interest -- fail to account for the heterogeneous dynamics of airport operations.
    While effective in autonomous driving, a fixed region in the aviation domain may either capture insufficient context for high-speed maneuvers (dark circles) or include an excessive number of irrelevant segments (bright circles), hindering context learning and increasing computational overhead.
    }
    \label{fig:teaser1}
    \vspace{-0.5cm}
\end{figure}

While prior work~\cite{navarro2024amelia} has shown that architectures developed for autonomous driving can theoretically be transferred to airport surface operations, there exist fundamental differences between road traffic and aviation.
Compared to car or pedestrian trajectory prediction, airport surface movements exhibit a substantially broader dynamic range.
Aircraft velocities at airports span from rather low speeds during taxiing to high speeds during takeoff and landing.
This is further amplified by rapid accelerations, \eg during a rolling start when an aircraft quickly gains speed.
As a result, motion dynamics are highly context-dependent and can change significantly within only a few seconds.
This dynamic variability leads to large differences in spatial extent within a fixed temporal horizon: depending on the operational phase, predicted trajectories may cover only a few meters or extend over several kilometers, \eg during takeoff.
Such variability poses a unique challenge for trajectory prediction models, which must simultaneously handle fine-grained low-speed maneuvers and long-range high-speed motion.
This stands in strong contrast to road traffic scenarios, where methods are designed to handle comparatively homogeneous motion patterns and scene scales.

We propose \mn~(\textbf{\underline{d}}irected \textbf{\underline{e}}dge \textbf{\underline{sc}}ene encoding for airport surface movem\textbf{\underline{ent}}), a trajectory prediction architecture tailored to the unique characteristics, especially heterogeneous dynamics, of airport surface movements.
By formulating the initial map topology as a graph of directed edges, we introduce a Potential Reachable Set–based sampling strategy to extract relevant airfield segment context while maintaining a manageable number of scene tokens.
To obtain relevant map elements, we traverse the lane topology graph and collect connected lane segments according to their semantic type. The traversal depth is determined by both the cumulative distance and the segment category; for structurally coherent elements such as runways, we include them in their entirety to preserve the overall context.
This design enables a sparse yet expressive representation, effectively capturing context for long-tail actions such as takeoffs without sacrificing critical contextual information.
Extensive experiments on the Amelia-10 benchmark demonstrate that our method achieves significant performance improvements over previous work.
Notably, it excels for agents annotated as safety-critical.
In summary, our main contributions include:
\begin{itemize}
    \item We introduce a potential-reachable-set-based scene context sampling to address the unique challenges of map context extraction in airport surface operations.
    \item We integrate this sampling strategy into a novel trajectory prediction model (\mn), achieving state-of-the-art performance on the Amelia-10 benchmark.
    \item Comprehensive evaluations on Amelia-10 demonstrate the benefits of our approach, particularly for safety-critical agents, while maintaining low inference latency and effectively capturing long-horizon scene context.

\end{itemize}

\section{RELATED WORK}

\subsection{Trajectory Prediction for Self-Driving}
Advances in autonomous driving facilitated significant research interest motion forecasting for road traffic~\cite{liu2021multimodal, nayakanti2023wayformer, zhou2023query, wang2023prophnet, cheng2023forecast, zhou2024smartrefine, tang2024hpnet, shi2024mtr++, song2024realmotion, zhang2024demo, huang2025trajectory, wan2025multi, zhou2025modeseq}.
The common high-level architecture first encodes agent and map data separately, then a scene encoder is used to model interactions. 
Map context is typically provided as a vector map, from which relevant elements are selected by sampling a fixed spatial region around the agent; these elements are then encoded using PointNet-like modules~\cite{qi2017pointnet}.
Early methods operate directly on lane graphs: LaneGCN~\cite{liang2020learning} encodes lanes using a graph neural network (GNN), while PGP~\cite{deo2022multimodal} explicitly learns policies for lane graph traversal.
Agent motion histories are encoded with self-attention~\cite{cheng2023forecast, prutsch24efficient, song2024realmotion} or state-space models~\cite{zhang2024demo, huang2025trajectory}.
Scene-level interactions are modeled either with GNNs~\cite{zhou2023query, jia2023hdgt, cui2023gorela} or attention-based modules~\cite{liu2021multimodal, shi2022motion, nayakanti2023wayformer}.
To generate multimodal output hypotheses, approaches~\cite{shi2022motion, nayakanti2023wayformer} commonly adopt detection-transformer-style~\cite{carion2020end} decoders.

\begin{figure*}[t]
    \centering
    \vspace{0.17cm}
    \includegraphics[trim={0cm, 0cm, 0cm, 0cm}, clip, width=0.99\linewidth]{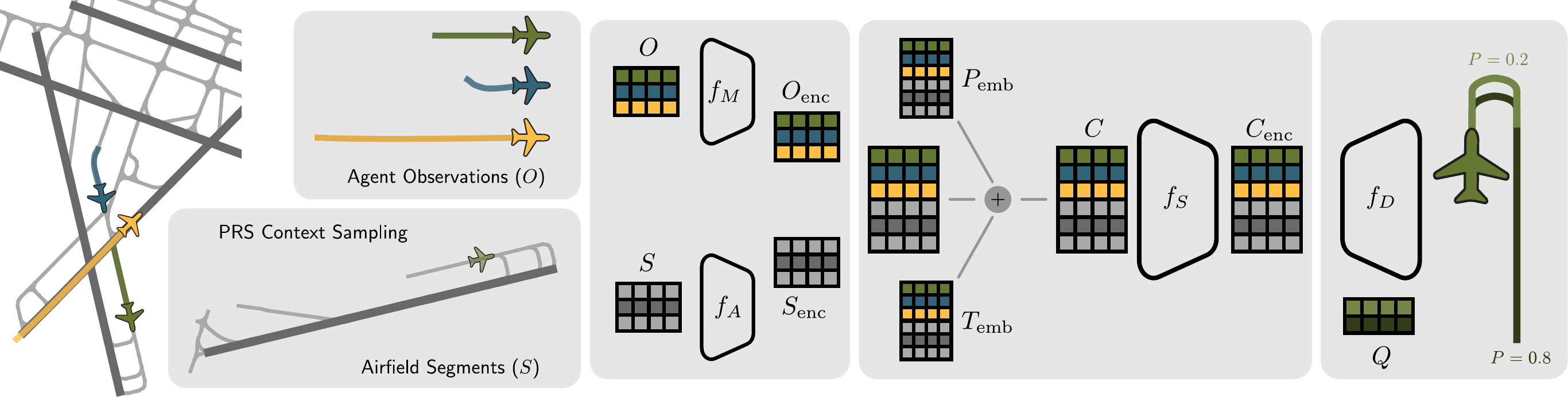}
    \vspace{-0.25cm}
    \caption{
    Overview of our \mn~architecture.
    We employ a potential reachable set-based airfield map sampling to obtain the map context based on the current focal agent position.
    Then, all agent observations $O$ and airfield segments $S$ are encoded using individual encoders $f_M$ and $f_A$.
    Positional $P_\text{emb}$ and type $T_\text{emb}$  embeddings are added to the tokens, to form a scene context $C$.
    The scene context is encoded using a scene encoder $f_S$, lastly a decoder using learnable queries $Q$ create output trajectories $T$ and probability scores $P$.
    }
    \label{fig:arch}
    \vspace{-0.5cm}
\end{figure*}

Research on motion forecasting for autonomous driving has led to sophisticated model components and architectures, providing also a strong foundation for other domains.
However, specialized use-cases, such as airport surface operations, pose unique challenges that limit the effectiveness of standard methods.
For instance, during takeoff or landing, an aircraft may traverse the entire airport, requiring alternative map-sampling strategies to capture relevant spatial context.

\subsection{Aircraft Trajectory Prediction}
The Amelia framework~\cite{navarro2024amelia} introduces a large-scale benchmark, Amelia-10, together with a baseline model, Amelia-TF, for airport surface movement prediction.
The Amelia-10 benchmark contains selected data collected from ten U.S. airports.
Its baseline model, Amelia-TF, builds upon SceneTransformer~\cite{ngiam2021scene}, an architecture originally proposed for autonomous driving.
Historical agent states are first projected into an embedding space using a multi-layer perceptron (MLP).
Map context is incorporated by sampling a fixed number of nearby map points for each agent and encoding them with a VectorNet-based encoder~\cite{gao2020vectornet}.
Agent interactions are modeled via factorized attention, decomposed into temporal self-attention, agent-to-map attention, and agent-to-agent attention.
The model predicts multiple plausible future trajectories parameterized as Gaussian Mixture Models (GMMs).
While Amelia-TF establishes a strong baseline, its map representation is limited to locally sampled geometric points.
This design neither adequately captures the broad dynamic range of aircraft motions nor explicitly models the structured topology of airport surface layouts.
Moreover, its reliance on SceneTransformer does not incorporate more recent architectural advances from self-driving research, which have achieved substantial performance gains on autonomous driving benchmarks~\cite{chang2019argoverse, ettinger2021large}.

Further related work on aircraft trajectory prediction~\cite{navarro2022social, patrikar2022trajair, yin2023context, yin2025aircraft, yang2025goodflight} primarily focuses on airborne scenarios, including commercial aviation and general aviation in non-towered airspace.
In airspace, agent interactions are sparse, motion is governed by continuous flight dynamics, and map constraints are only flight sector-based and comparatively weak.
In contrast, airport surface operations involve dense traffic, frequent interaction, discrete routing structures, and strong geometric and operational constraints.
Due to these fundamental differences in interaction density, motion constraints, and environmental structure, models developed for airborne trajectory prediction are not directly applicable to airport surface movement prediction.
Taxiing aircraft operate within highly structured, topology-constrained environments and exhibit motion characteristics that are more akin to constrained ground vehicles than to airborne aircraft.

\section{Airport Surface Movement Prediction using Directed Edge Scene Encoding}
This section details the architecture of \mbox{\mn}~(see Figure~\ref{fig:arch}).
Following the airport surface trajectory prediction task defined in the Amelia framework~\cite{navarro2024amelia}, our model takes historical observations over a past horizon of $H_p$ time steps and airfield map data as input.
The objective is to predict $k$ future trajectory hypotheses and associated confidence scores for a focal agent over a future horizon $H_f$.
By generating multimodal predictions, our model accounts for the inherent uncertainty in taxiing maneuvers, ensuring high coverage of potential future actions.
The core contribution of our approach is a novel domain-aware Potential Reachable Set (PRS) scene sampling mechanism. This technique provides physically grounded context to an efficient transformer-based backbone, allowing the network to focus on feasible maneuvers within the constrained airfield environment.

\begin{figure}[t]
    \centering
    \vspace{0.17cm}
    \newsavebox{\rightimg}
    \savebox{\rightimg}{\includegraphics[trim={0cm, 0cm, 0cm, 0cm}, clip, width=0.6\linewidth]{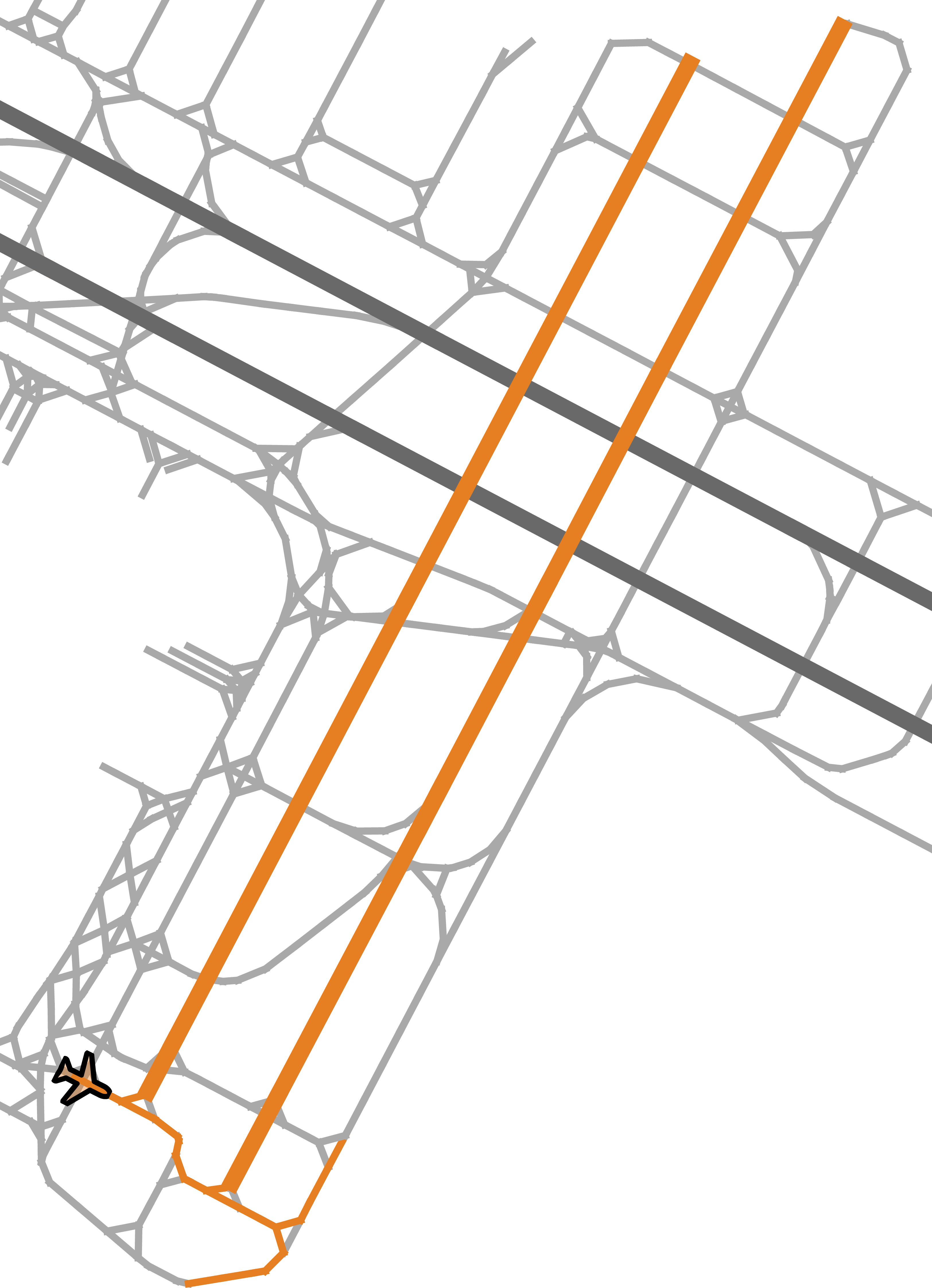}}

    \begin{minipage}[b][\ht\rightimg][s]{0.35\linewidth}
        \centering
        \includegraphics[trim={0cm, 0cm, 0cm, 0cm}, clip, width=\linewidth]{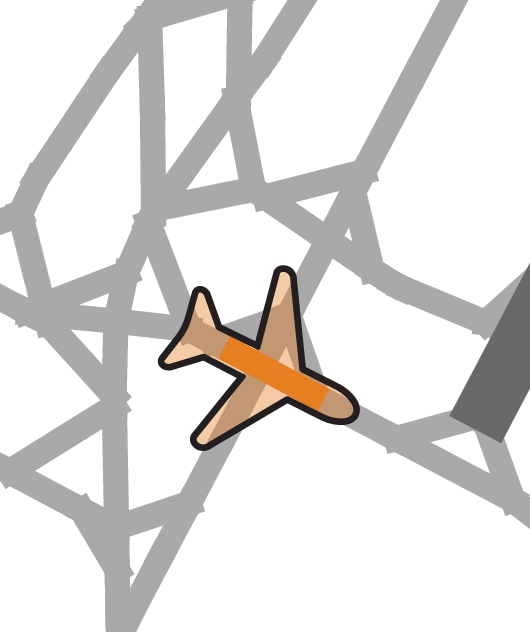}
        \vfill %
        \includegraphics[trim={0cm, 0cm, 0cm, 0cm}, clip, width=\linewidth]{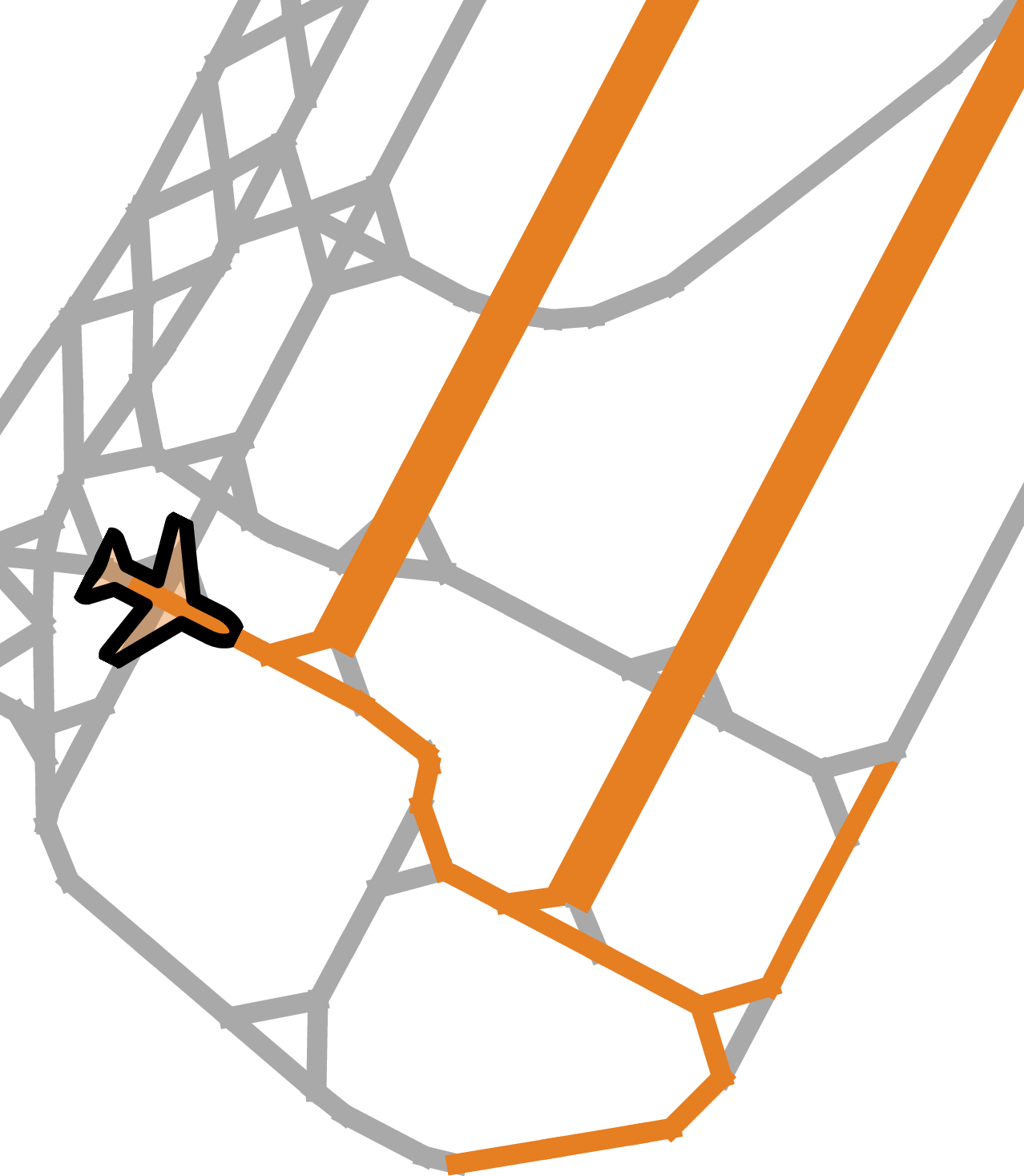
        }
    \end{minipage}%
    \hfill 
    \begin{minipage}[b]{0.6\linewidth}
        \usebox{\rightimg}
    \end{minipage} \
    \vspace{-0.25cm}
    \caption{
        Illustration of our Potential Reachable Set-based context sampling (on the KSFO airport).
        The process begins by locating the agent’s current airfield segment (top left), followed by a graph traversal to identify reachable pathways (bottom left).
        We generate a sparse context (right) comprising local segments and distant but reachable long-term structures.
        This selection reflects the agent’s operational constraints, prioritizing high-velocity paths like runways over distant, unreachable taxiways.
    }
    \vspace{-0.5cm}
    \label{fig:prs}
\end{figure}

\subsection{Airfield Segment Sampling}
To account for the unique characteristics of airport surface movements--where trajectories within a fixed time horizon may span only a few meters for ground vehicles and taxiing aircraft, yet extend several kilometers for aircraft during takeoff and landing--we propose a domain-aware environment sampling strategy.
Specifically, we employ a Potential Reachable Set-based approach that captures relevant context along all feasible paths and provides it to the model as a structured input.
We define the Potential Reachable Set (PRS) as the set of all airfield segments accessible from a agent's current location via a valid path in the lane graph.
Unlike standard methods that rely on fixed-size regions of interest~\cite{cheng2023forecast} or a predefined number of nearest tokens~\cite{zhou2023query, navarro2024amelia}, our approach adaptively selects context based on infrastructure constraints.
This keeps the input token counts manageable while preserving all information relevant to trajectory prediction.

The reachable distance within the PRS is bounded using lane-type-specific thresholds rather than instantaneous vehicle metrics. Current velocity is often a poor predictor of path length within a future prediction frame; for instance, an aircraft stationary at a runway threshold will likely cover a significantly larger distance over the next horizon than one taxiing toward a terminal, despite having a lower initial velocity.
To implement this, we first reorganize raw map data into semantically meaningful airfield segments as a preprocessing step.
We merge consecutive graph edges belonging to the same structural element (\eg a specific runway or taxiway) while preserving intersection nodes and edge directions to maintain topological consistency.
We enforce a maximum segment length based on the element type to ensure balanced granularity, thereby reducing graph complexity without sacrificing structural detail.

Figure~\ref{fig:prs} outlines the process of our PRS-based sampling.
To extract the environmental context for a given vehicle, we first identify its current segment.
We select potential candidates based on Euclidean distance, refined by a heading consistency constraint to avoid incorrect associations in dense intersection areas.
If the closest segment’s orientation deviates significantly from the vehicle's heading, we evaluate the next $n$ candidates, defaulting to the closest segment only if no angular match is found within the set.
Once the current segment is detected, the lane graph allows for the inference of multiple feasible future paths.
To enable efficient model inference, we precompute all admissible paths originating from each segment and store the segment IDs as metadata.
This is achieved via a depth-first traversal of the lane graph, constrained by domain-specific length thresholds.
For example, if a runway segment is involved, the entire runway is included to support full coverage of long-horizon takeoff and landing trajectories.
This results in a memory-efficient representation that allows the map to be filtered during test time with minimal computational overhead.

\subsection{Airfield Segment Encoding}
Building upon our airfield sampling, we encode the segments $S \in \mathbb{R}^{N_s \times P_s \times 2}$, where $N_s$ is the number of segments, $P_s$ is the number of 2D points sampled per segment, using a PointNet-like~\cite{qi2017pointnet} airfield encoder $f_A$, following common practice from autonomous driving~\cite{cheng2023forecast}.
Before applying the encoder, each lane segment is transformed to a segment-centric local coordinate system by normalizing it based on its center pose.
This facilitates more efficient learning of lane shapes independent of their global positions~\cite{cheng2023forecast}.
As a result, we obtain one token per segment, leading to a map context $S_\text{enc} \in \mathbb{R}^{N_s \times D}$, where $D$ is the model feature dimension.

\subsection{Motion Encoding}
The historical agent observations are represented in a tensor $O \in \mathbb{R}^{N_a \times T_h \times D_m}$, where $N_a$ denotes the number of agents, $T_h$ the number of historical time steps, and $D_m$ the input feature dimension, including, \eg 3D positions and heading.
We employ a lightweight yet effective self-attention–based motion encoder $f_M$~\cite{prutsch24efficient} to obtain motion tokens $O_\text{enc} \in \mathbb{R}^{N_a \times D}$.
All observations are transformed into a local coordinate frame defined by the most recent aircraft pose.
This normalization allows the encoder to focus on motion patterns independent of global position and orientation.
First, the input observations are projected into the model feature space via a linear projection layer.
Subsequently, self-attention~\cite{vaswani2017attention} is applied along the temporal dimension, followed by pooling over time to aggregate the historical information into a compact per-agent representation.

\subsection{Scene Encoding and Trajectory Decoding}
Following the initial local encoding of agent observations and airfield segments, we construct a scene context $C \in \mathbb{R}^{(N_a+N_s) \times D}$ by concatenating agent and map tokens.
Each token is augmented with a positional embedding $P_\text{emb}$ based on the pose of its respective local coordinate system~\cite{cheng2023forecast}, \ie the segment center for map elements and the most recent pose for agents.
This embedding is implemented as a two-layer MLP that encodes 3D position and yaw.
Additionally, we incorporate categorical embeddings $T_\text{emb}$ to distinguish between airfield segment types and agent classes~\cite{cheng2023forecast}.
An attention-based scene encoder~$f_S$ then models the relational dependencies between the focal agent and all other scene elements, yielding an updated context $C_\text{enc} \in \mathbb{R}^{(N_a+N_s) \times D}$.

To generate multimodal trajectory predictions, we adopt a decoder $f_D$ inspired by detection transformers~\cite{carion2020end}.
We use learnable mode queries $Q \in \mathbb{R}^{k \times D}$, where $k$ is the number of intention modes, and apply factorized cross-attention over the encoded scene context.
The updated mode tokens $Q'$ parameterize a Gaussian Mixture Model (GMM) representing the future distribution.
Finally, shallow MLP heads map these tokens to the predicted trajectory means $M \in \mathbb{R}^{k \times T_f \times 3}$, variances $V \in \mathbb{R}^{k \times T_f \times 3}$, and confidence scores $P \in \mathbb{R}^k$, where $T_f$ is the future temporal horizon.

\section{EXPERIMENTAL SETUP}

\subsection{Datasets and Data Preprocessing}
\begin{figure}[t]
    \centering
    \vspace{0.17cm}
    \includegraphics[trim={0cm, 0cm, 0cm, 0cm}, clip, width=1\linewidth]{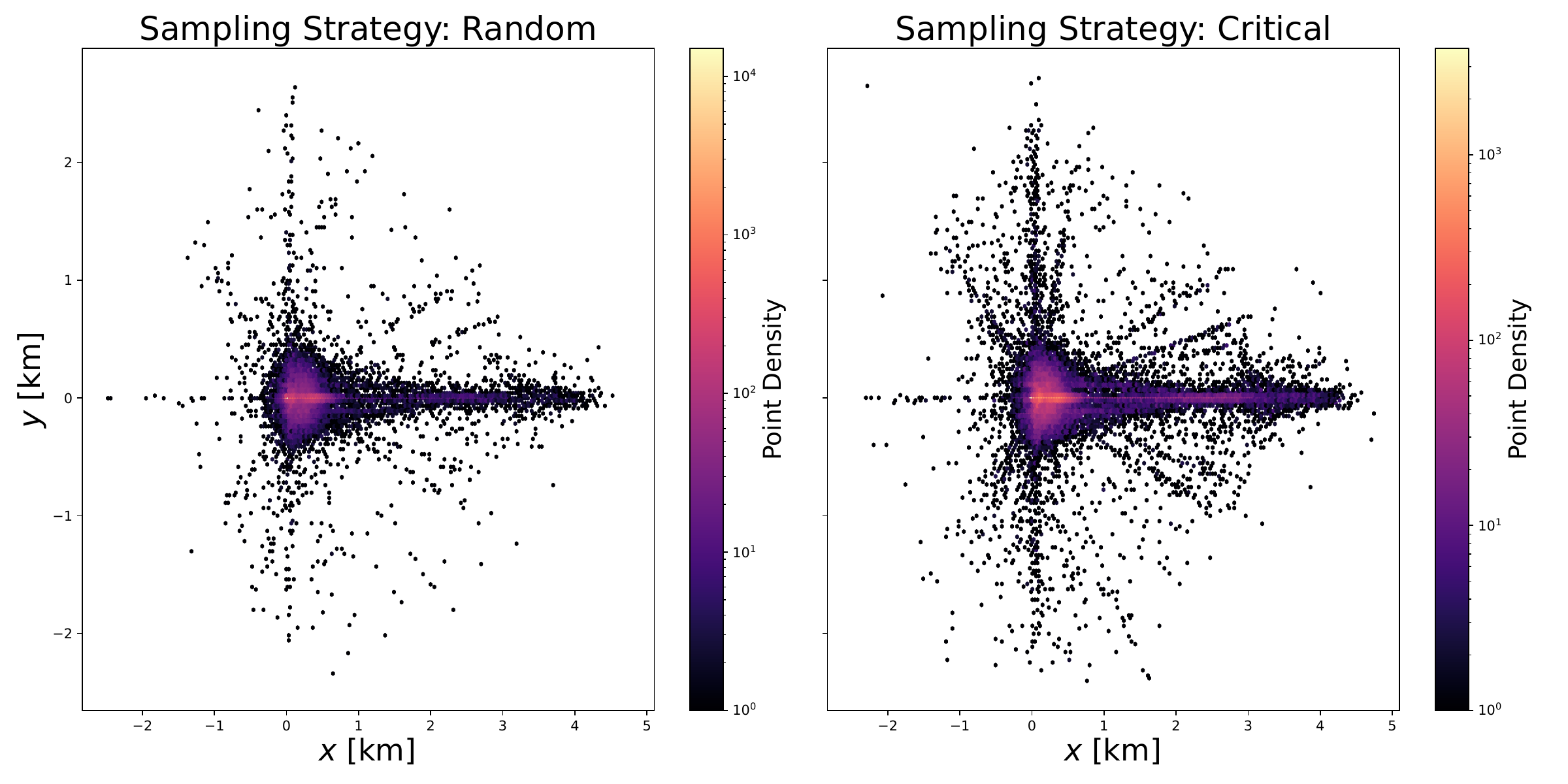}
    \includegraphics[trim={0cm, 0cm, 0cm, 0cm}, clip, width=1\linewidth]{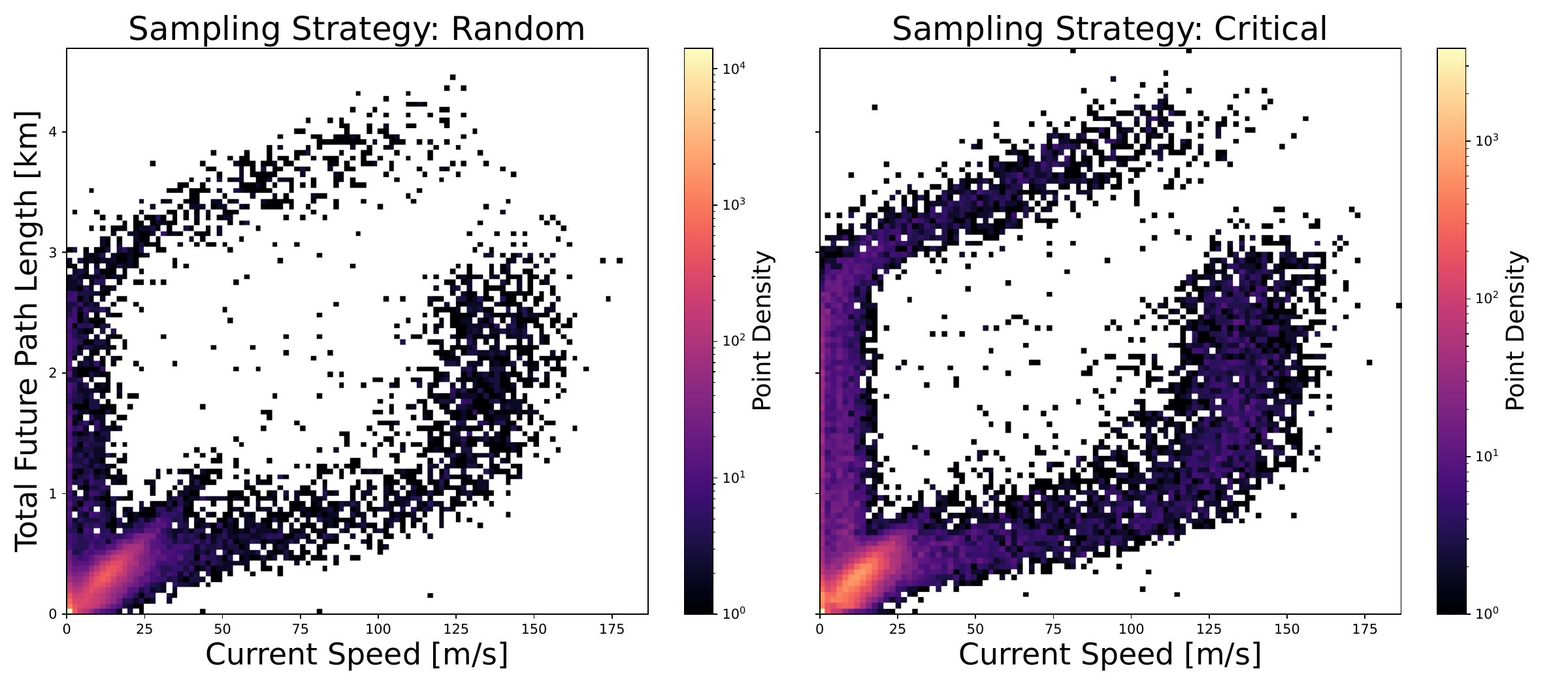}
    \vspace{-0.3cm}
    \caption{
    Comparison of random vs. critical focal agent sampling in the Amelia-10 benchmark.
    The top row displays the distribution of future trajectory endpoints, highlighting a more diverse distribution for critical agents.
    The bottom row illustrates the relationship between current agent speed and future trajectory length.
    Notably, critical agents show higher dynamic changes: agents with low initial speed but long future trajectories (acceleration) and high initial speed with short future paths (braking)--compared to the more linear correlation seen in random agents.
    Best viewed digitally with zoom.
    }
    \vspace{-0.5cm}
    \label{fig:dataset}
\end{figure}

We train and evaluate our model on the Amelia-10 benchmark~\cite{navarro2024amelia}, a large-scale dataset for aircraft surface movement trajectory prediction.
The dataset contains recordings from ten U.S. airports: KBOS, KDCA, KEWR, KJFK, KLAX, KMDW, KMSY, KSEA, KSFO, and PANC.
The airports show different characteristics such as varying number of runways and runway layouts, terminal configurations, and traffic densities. 
For preprocessing, we use the official implementation and configuration provided by the dataset authors\footnote{\url{https://github.com/AmeliaCMU/AmeliaScenes}}.
The preprocessing pipeline generates dedicated training, validation, and test splits for each airport.
Each scenario may contain multiple agents and the framework includes several strategies to select a focal agent.
We report results for two evaluation setups: the baseline approach where the focal agent is selected randomly, and a criticality-based selection strategy that leverages the criticality score provided in the dataset~\cite{navarro2024amelia}.
The criticality score prioritizes agents near potential conflict points while down-weighting stationary agents.
Consequently, the evaluation set under critical-agent selection exhibits substantially higher variability and interaction complexity.
Figure~\ref{fig:dataset} compares the data distribution for random and critical focal agent sampling.
During training, we employ random focal agent selection.
Following the benchmark protocol, we use a sampling rate of 1\,Hz, a past observation length of $H_p=10\,s$ resulting in $T_h=10$ input steps; prediction horizon of $H_f=50\,s$ ($T_f=50$ output steps).
We set the number of trajectory modes $k=4$ to align with the evaluation protocol.
To maintain reasonable training times, we train and validate on half of the dataset, reporting results on the complete test set.

\subsection{Implementation Details}
We use a latent dimension of $D=128$.
Based on empirical validation, the encoders $f_M$ and $f_S$ each consist of four attention blocks, while the decoder $f_D$ utilizes a cross-attention depth of three.
Each (GMM) output head is implemented as a multi-layer perceptron (MLP) with hidden dimensions $D$ and $2D$, separated by a ReLU activation function.
The implementations of the airfield encoder $f_A$ and positional embeddings follow~\cite{cheng2023forecast}.
We utilize the 3D position, speed, and yaw as agent input features ($D_a=5$) and sample $P_s=20$ points per airfield segment.
Our model distinguishes between two agent classes (ground vehicles and aircraft) and three airfield segment types (taxiways, runways, and hold lines).
During graph preprocessing, the maximum segment length is constrained to 0.8$\,$km for runways and 30$\,$m for all other segments to ensure balanced granularity. For the PRS-based context sampling, we set the maximum path depth to 20 segments, corresponding to a maximum 600$\,$m taxiway horizon.
Furthermore, we perform a backward traversal from the focal agent's current position to incorporate historical path context, providing a more comprehensive representation of the local scene topology.

Due to the varying number of scenarios per airport, we adapt the number of training epochs individually, terminating once convergence on the validation set is observed.
We use a batch size of 128 and initialize the learning rate at $10^{-4}$.
When the validation loss plateaus, the learning rate is reduced by one order of magnitude; training concludes once a learning rate of $10^{-8}$ is reached.
Training is done on a single NVIDIA GPU with 48\,GB VRAM.
To optimize our model, we employ a winner-takes-all objective--a standard strategy in multi-modal trajectory prediction in which only the hypothesis with the smallest displacement error contributes to the loss.
The overall objective consists of three components.
For mode selection, a standard cross-entropy classification loss encourages the highest confidence for the best-fitting trajectory.
For trajectory regression, we use a Negative Log-Likelihood (NLL) loss parameterizing the predicted distribution via the mean and variance.

In addition, we introduce a signed distance field (SDF)-based regularization term to enforce map compliance.
Given that airport surface motion is constrained to predefined segments, predicted trajectories must remain consistent with the underlying map topology.
The SDF loss penalizes spatial deviations from these valid areas, thereby promoting topologically feasible trajectories.
To compute this term, we precompute a rasterized SDF at a 5\,m resolution, where each cell stores the Euclidean distance to the nearest airfield segment.
During training, predicted trajectory points are projected onto the grid, the corresponding distance values are queried, averaged across all time steps, and incorporated into the overall loss to penalize off-track predictions.

\begin{table*}[t]
\setlength{\tabcolsep}{5pt}
    \vspace{0.17cm}
    \caption{
    Trajectory prediction results on the Amelia-10-bench test set for forecasting horizons of 50\,s and 20\,s.
    Experiments are done on the test sets, evaluating on the most \emph{critical} agent per scenario.
    Errors reported in meters.
    }
    \vspace{-0.4cm}
    \label{tab:res_single_crit}
    \begin{center}
    \begin{tabular}{l|cccccccccc|c|c}
    Method            & {KMDW}     & {KEWR}     & {KBOS}     & {KSFO}*     & {KSEA}*     & {KDCA}     & {PANC}     & {KLAX}     & {KMSY}     & {KJFK}     & Avg\textsuperscript{\dag} & Metric \\ \hline
    Amelia-TF~\cite{navarro2024amelia}     & 66.69 & 91.18 & 81.61 &     - &     - & 105.60 & 170.48 & 137.40 & 87.46 & 102.77 & 105.40 & \multirow{2}{*}{mFDE@50s} \\
    \ourrow                                & 59.64 & 77.74 & 64.49 & 56.60 & 61.45 & \phantom{0}54.04 & \phantom{0}67.13 & \phantom{0}64.50 & \phantom{0}68.98 & \phantom{0}78.67 & \phantom{0}66.90                           \\ \arrayrulecolor{lightgray}\hline\arrayrulecolor{black}
    Amelia-TF~\cite{navarro2024amelia}     & 26.41 & 32.88 & 30.79 &     - &     - & \phantom{0}39.24 & \phantom{0}61.35 & \phantom{0}50.74 & \phantom{0}31.41 & \phantom{0}39.23 & \phantom{0}39.01 & \multirow{2}{*}{mADE@50s} \\
    \ourrow                                & 24.88 & 31.69 & 26.80 & 23.22 & 24.79 & \phantom{0}25.46 & \phantom{0}28.41 & \phantom{0}27.43 & \phantom{0}27.39 & \phantom{0}33.56 & \phantom{0}28.20                           \\ \hline
    Amelia-TF~\cite{navarro2024amelia}     & 13.35 & 15.78 & 14.74 &     - &     - & \phantom{0}20.20 & \phantom{0}28.08 & \phantom{0}25.79 & \phantom{0}14.65 & \phantom{0}19.82 & \phantom{0}19.05 & \multirow{2}{*}{mFDE@20s} \\
    \ourrow                                & 13.60 & 17.20 & 14.25 & 12.40 & 12.68 & \phantom{0}15.76 & \phantom{0}15.44 & \phantom{0}15.21 & \phantom{0}14.88 & \phantom{0}19.29 & \phantom{0}15.70                           \\ \arrayrulecolor{lightgray}\hline\arrayrulecolor{black}
    Amelia-TF~\cite{navarro2024amelia}     & \phantom{0}6.94 &  \phantom{0}7.79 &  \phantom{0}7.33 &     - &     - & \phantom{0}10.26 & \phantom{0}14.55 & \phantom{0}12.15 &  \phantom{00}7.36 &  \phantom{00}9.65 &  \phantom{00}9.50 & \multirow{2}{*}{mADE@20s} \\
    \ourrow                                &  \phantom{0}7.23 &  \phantom{0}8.86 &  \phantom{0}7.32 & \phantom{0}6.49 &  \phantom{0}6.64 &  \phantom{00}8.63 &  \phantom{00}8.06 &  \phantom{00}8.21 &  \phantom{00}7.86 & \phantom{00}9.80 &  \phantom{00}8.25                           \\
    \end{tabular} \\
    \vspace{0.1cm}
    *No checkpoints available for the official, refactored codebase. \dag~Excluding KSFO and KSEA.
    \end{center}
    \vspace{-0.5cm}
\setlength{\tabcolsep}{4pt}
\end{table*}

\begin{table*}[t]
\setlength{\tabcolsep}{5pt}
    \vspace{0.17cm}
    \caption{
    Trajectory prediction results on the Amelia-10-bench test set for a forecasting horizon of 50\,s.
    Experiments are done on the test sets, evaluating a \emph{random} agent per scenario.
    Errors reported in meters.
    }
    \vspace{-0.4cm}
    \label{tab:res_single_rand}
    \begin{center}
    \begin{tabular}{l|cccccccccc|c|c}
    Method            & {KMDW}     & {KEWR}     & {KBOS}     & {KSFO}*     & {KSEA}*     & {KDCA}     & {PANC}     & {KLAX}     & {KMSY}     & {KJFK}     & Avg & Metric \\ \hline
    Amelia-TF~\cite{navarro2024amelia}     & 27.52 & 52.06 & 51.01 &     40.23 &    65.82 & 47.75 & 86.24 & 89.46 & 29.84 & 55.32 & 54.52 & \multirow{2}{*}{mFDE@50s} \\
    \ourrow                                & 25.23 & 48.80 & 42.23 & 37.11 & 39.56 & 27.92 & 44.14 & 45.17 & 24.63 & 46.27 & 38.11                           \\ \arrayrulecolor{lightgray}\hline\arrayrulecolor{black}
    Amelia-TF~\cite{navarro2024amelia}     & 11.80 & 20.96 & 20.47 &     17.05 &     29.94 & 18.93 & 34.90 & 35.84 & 11.42 & 22.43 & 22.37 & \multirow{2}{*}{mADE@50s} \\
    \ourrow                                & 11.37 & 21.34 & 18.35 & 15.81 & 17.51 & 13.21 & 20.34 & 20.24 & 10.38 & 20.51 & 16.91                           \\
    \end{tabular} \\
    \vspace{0.1cm}
    * No checkpoints available for the official, refactored codebase. Values taken from reported results~\cite{navarro2024amelia}.
    \end{center}

        \vspace{-0.5cm}
\setlength{\tabcolsep}{4pt}
\end{table*}

\subsection{Metrics}
We evaluate our models using the standard metrics for the Amelia-10 benchmark: minimum Average Displacement Error (mADE) and minimum Final Displacement Error (mFDE).
Both metrics are computed by selecting, from the set of $k$ predicted hypotheses, the trajectory that best matches the ground truth.
The mADE measures the mean Euclidean ($L^2$) distance between predicted and ground-truth positions across all future time steps, whereas the mFDE considers only the endpoints.
Following the benchmark protocol, we report results for two prediction horizons: 20\,s and 50\,s.

\section{RESULTS AND DISCUSSION}

We report results for single-airport experiments, assessing performance on both safety-critical agents and randomly selected focal agents.
As a baseline, we re-run the Amelia-TF model using the official codebase and pretrained weights\footnote{\url{https://huggingface.co/AmeliaCMU/AmeliaTF-weights-only}}, since the framework has undergone several refactoring and the currently reproducible results differ from those reported in the original paper.
This ensures a fair comparison on identical data splits.
In addition to the single-airport setting, we provide a cross-airport evaluation, an ablation study of our proposed context sampling strategy, and a detailed latency analysis.
We restrict our comparison to Amelia-TF as a baseline, since adapting methods designed for autonomous driving to Amelia-10 would require substantial architectural modifications (\eg input dimensionality, lane representation, and sampling strategies), allowing implementation choices to heavily influence performance and thereby limiting the fairness of the comparison.

\begin{table*}[t]
\setlength{\tabcolsep}{5pt}
    \vspace{0.17cm}
    \caption{
    Multi-airport trajectory prediction results on the Amelia-10-bench test set for a forecasting horizon of 50\,s.
    Experiments are done on the test sets, evaluating a \emph{random} agent per scenario.
    Errors reported in meters.
    }
    \vspace{-0.4cm}
    \label{tab:multi_airport}
    \begin{center}
    \begin{tabular}{l|cccccccccc|c|c}
    Method            & {KMDW}     & {KEWR}     & {KBOS}     & {KSFO}     & {KSEA}     & {KDCA}     & {PANC}     & {KLAX}     & {KMSY}     & {KJFK}     & Avg & Metric \\ \hline
    \ourrow & 29.73 & 50.37 & 47.94 & 44.23 & 46.50 & 29.99 & 53.46 & 51.08 & 23.31 & 41.48 & 41.81 & mFDE@50s \\ \arrayrulecolor{lightgray}\hline\arrayrulecolor{black}
    \ourrow & 13.01 & 21.57 & 20.39 & 18.83 & 20.35 & 14.01 & 24.08 & 22.75 & 10.10 & 17.85 & 18.28 & mADE@50s \\
    \end{tabular} \\
    \vspace{0.1cm}
    \end{center}

    \vspace{-0.5cm}
\setlength{\tabcolsep}{4pt}
\end{table*}

\subsection{Single Airport Evaluation}
\begin{figure*}[t]
    \centering
    \vspace{0.17cm}
    \includegraphics[trim={0cm, 0cm, 0cm, 0cm}, clip, width=0.32\linewidth]{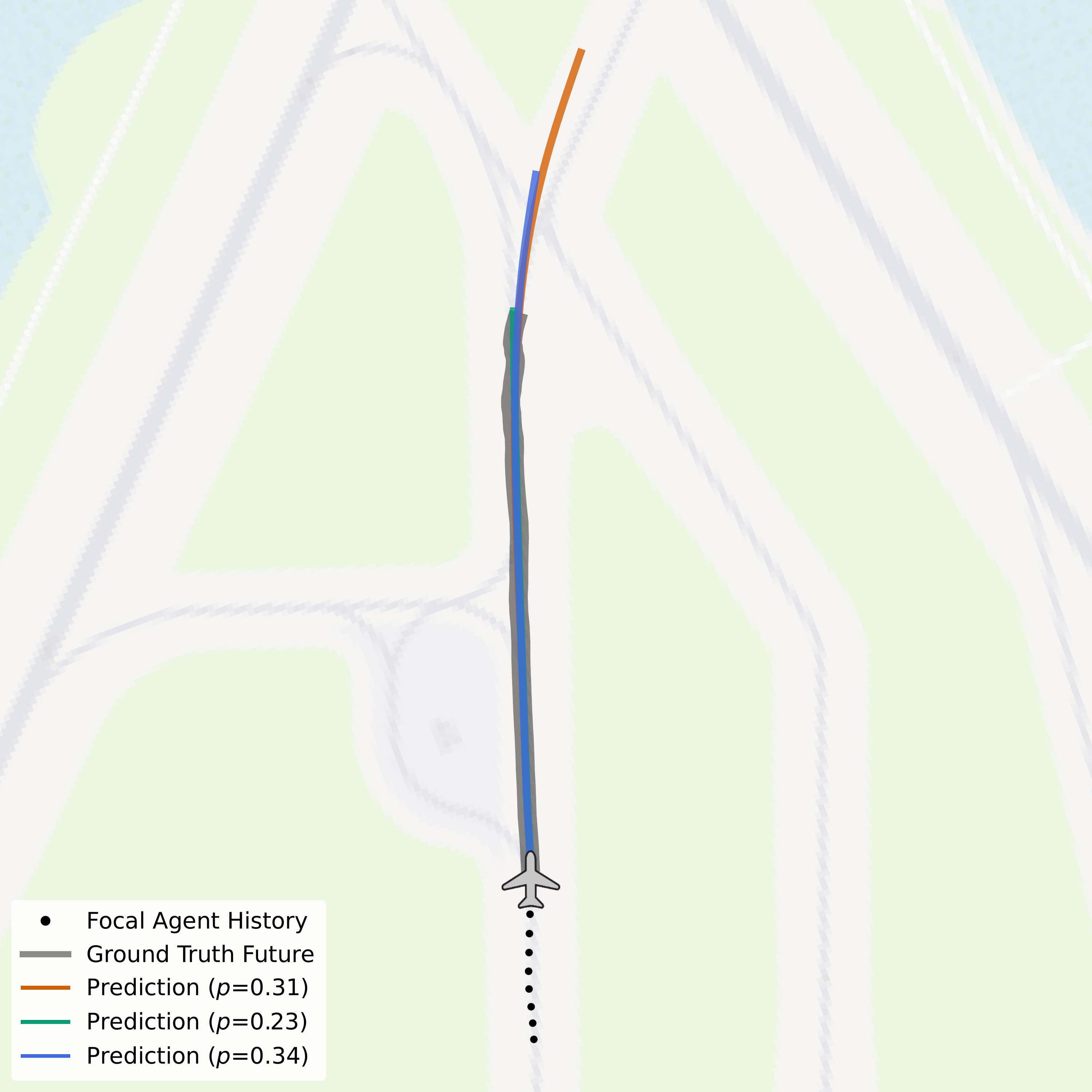}
    \includegraphics[trim={0cm, 0cm, 0cm, 0cm}, clip, width=0.32\linewidth]{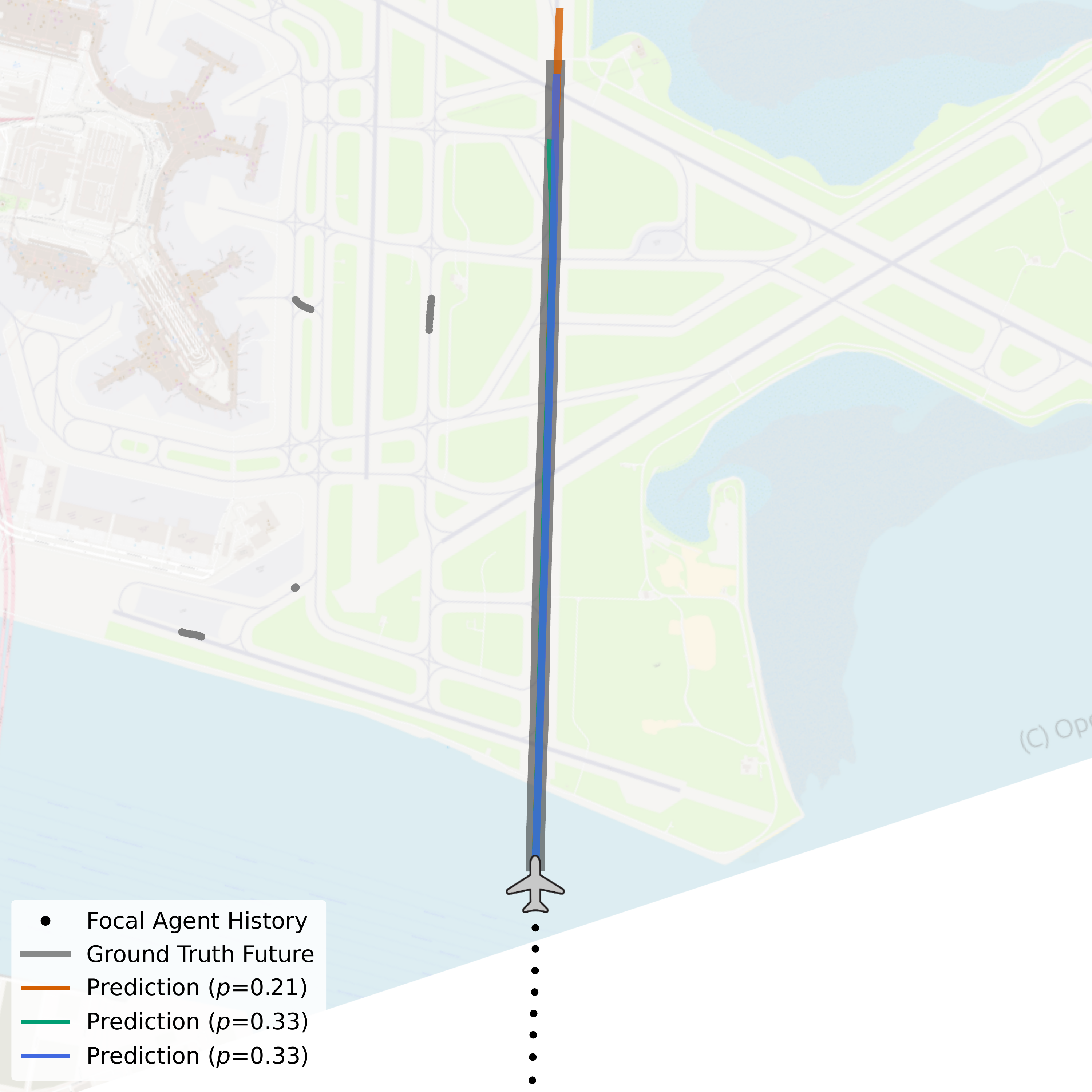}
    \includegraphics[trim={0cm, 0cm, 0cm, 0cm}, clip, width=0.32\linewidth]{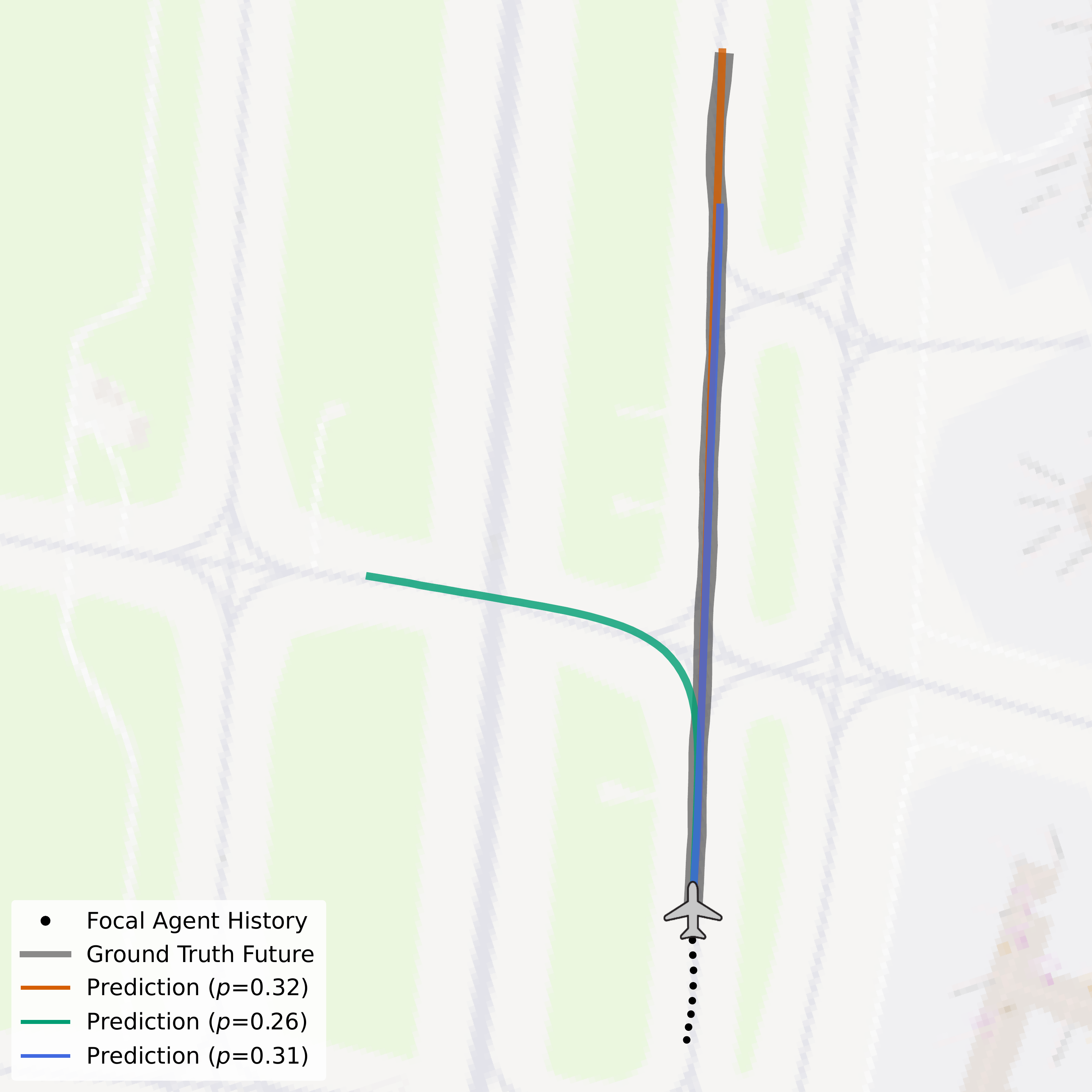}
    \medskip
    
    \includegraphics[trim={0cm, 0cm, 0cm, 0cm}, clip, width=0.32\linewidth]{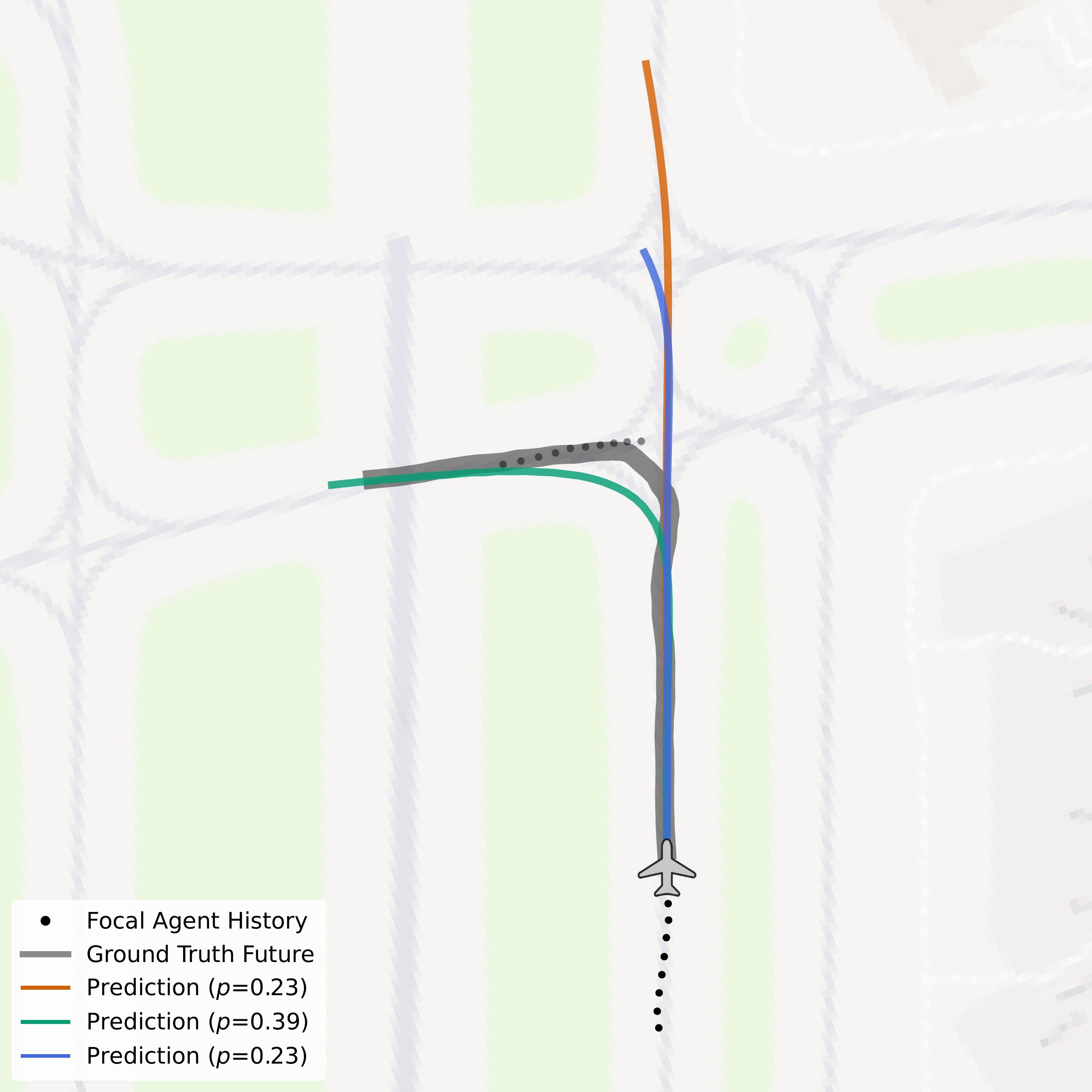}
    \includegraphics[trim={0cm, 0cm, 0cm, 0cm}, clip, width=0.32\linewidth]{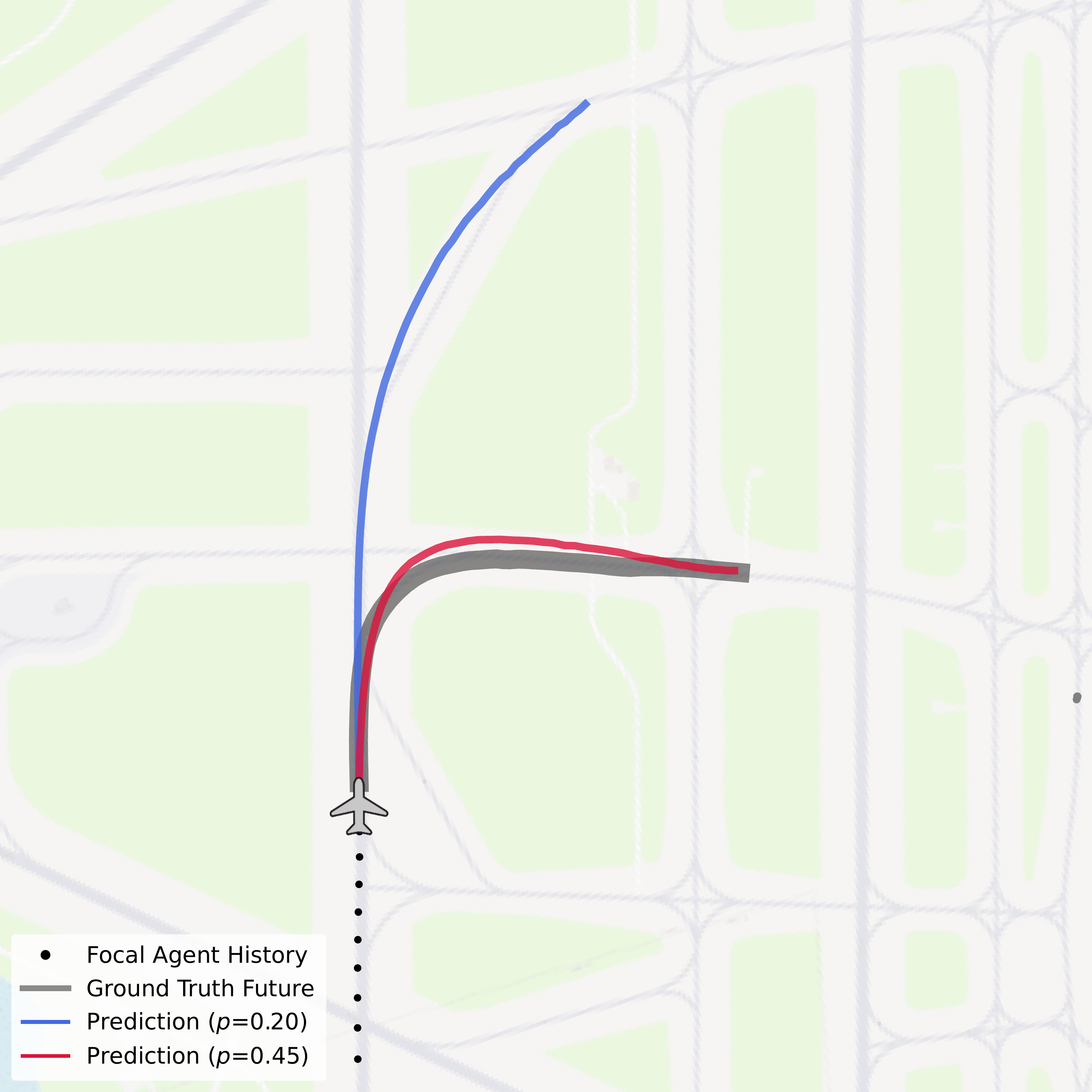}
    \includegraphics[trim={0cm, 0cm, 0cm, 0cm}, clip, width=0.32\linewidth]{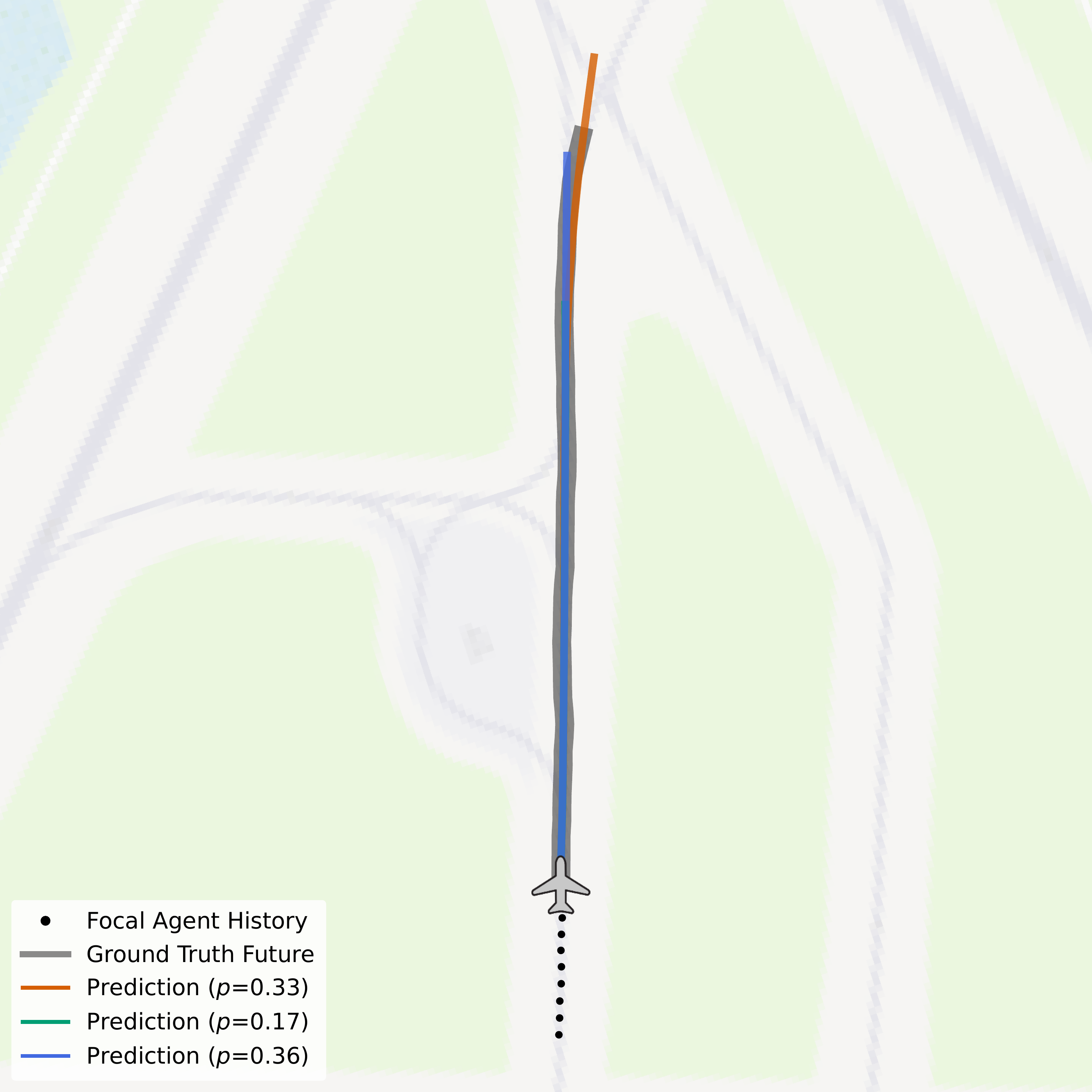}
    \vspace{-0.2cm}
    \caption{
    We visualize $50\,\text{s}$ \mn~predictions (colored lines) alongside the $10\,\text{s}$ historical focal agent movement (black), surrounding agent histories (light gray), the ground truth future trajectory (thick, dark gray), and airfield map context.
    For clarity, predictions with confidence scores below 0.15 are omitted.
    The scenarios are sampled from the Amelia-10 test set showing scenes from KBOS.
    Best viewed digitally with zoom.
    \vspace{-0.5cm}
    }
    \label{fig:results}
\end{figure*}

Table~\ref{tab:res_single_crit} presents the evaluation results on the critical-agent test setup.
The results show that in safety-critical scenarios and for a long prediction horizon of 50\,s our approach significantly outperforms the Amelia-TF baseline.
This underscores the effectiveness of our model in capturing long-range contextual dependencies and complex scene interactions.
For a shorter prediction horizon of 20\,s, the performance gap narrows and results are largely comparable, as shorter-term predictions are inherently less challenging.
Moreover, the benefit of incorporating sparse heterogeneous context structures is less pronounced in settings where agents operate primarily within a limited spatial vicinity around their current position.
The per-airport accuracies also correlate with the average spatial extent of the ground-truth trajectories.
For instance, future trajectories at KMDW average approximately 430\,m, whereas airports exhibiting higher absolute displacement errors, such as PANC and KLAX, have average trajectory lengths exceeding 700\,m.
This suggests that the error metrics are sensitive to the operational scale of the specific airfield, where higher taxiing and takeoff speeds naturally result in larger longitudinal deviations.
Figure~\ref{fig:results} illustrates qualitative trajectory prediction results generated by \mn, showcasing its ability to produce multimodal hypotheses that strictly adhere to the airfield topology across diverse operational phases.

\begin{table}[t]
\setlength{\tabcolsep}{4pt}
    \vspace{0.17cm}
    \caption{Ablation study on scene context sampling methods.
    We report the $mFDE@50s$ on the respective test sets using the most \emph{critical} agent per scenario.
    Errors reported in meters.}
    \vspace{-0.4cm}
    \label{tab:abl}
    \begin{center}
    \begin{tabular}{c|ccc|c}
    Scene Context Sampling & KBOS & KSFO & PANC & Avg \\ \hline 
    Top-K closest (50 segments) & 129.82 & 122.75 & 169.92 & 140.83 \\
    Radius-based (0.3 km) & \phantom{0}86.95 &  \phantom{0}80.44 & 107.69 &  \phantom{0}91.70 \\
    \rowcolor{rcol} PRS-based &  \phantom{0}64.49 &  \phantom{0}56.60 &  \phantom{0}67.13 &  \phantom{0}62.74 \\    
    \end{tabular}
    \end{center} 
    \vspace{-0.3cm}
\end{table}

Table~\ref{tab:res_single_rand} presents the motion prediction results for the random focal agent test setup.
In this configuration, the prediction task is less challenging (cf. Figure~\ref{fig:dataset}), resulting in generally lower overall prediction errors compared to the critical-agent setup.
Despite the reduced average difficulty, our approach continues to demonstrate significant performance advantages, particularly in long-term prediction as measured by mFDE. Regarding the mADE, our model achieves better or comparable results across all evaluated airports, further validating the robustness of the PRS-based context sampling.

\subsection{Multi Airport Evaluation}
We present a multi-airport evaluation in Table~\ref{tab:multi_airport}, where a single model is trained jointly on data from the Amelia-10 airports.
Comparing these results to the single-airport experiments (\cf Table~\ref{tab:res_single_rand}), the joint model actually surpasses the performance of the individual models at KJFK and KMSY, likely due to the increased diversity of training scenarios.
Overall, the unified model achieves high accuracy across all airports, nearly matching the performance of specialized, per-airport models.
Performance degradation is most pronounced at PANC and KLAX, two of the most challenging airports in the dataset, highlighting that for airfields with highly unique operational patterns, specialized models can be favorable.

\subsection{Ablation Study}
Table~\ref{tab:abl} compares our approach using different context sampling methods.
We test the standard approaches--using the top-$k$ closest lane segments or all segments within a predefined radius--to our Potential Reachable Set-based sampling.
The table reports mFDE results (critical agent sampling methods) for an easy airport (KSFO), a medium difficulty airport (KBOS) and the most challenging airport (PANC).
Across all airports, our approach achieves the best result, highlighting the effectiveness of our scene sampling.

\subsection{Latency Analysis}
Our approach offers a significant reduction in model complexity, featuring only 3.2M parameters compared to the 89.8M in Amelia-TF~\cite{navarro2024amelia}.
This is primarily attributed to a reduced latent dimension ($D=128$ vs. $D=256$) and a more compact map encoder architecture. 
On an NVIDIA L40 GPU, our model achieves inference latencies of 32\,ms, 68\,ms, and 132\,ms for predicting 32, 64, and 128 agents, respectively.
While this latency is higher than Amelia-TF (16\,ms, 36\,ms, 76\,ms for the respective batch sizes), the latter incorporates minimal map context.
Our method, in contrast, leverages a more comprehensive environmental representation, resulting in significantly improved accuracy.

The hyperparameters of our PRS implementation significantly impact runtime, as they determine the number of airfield segments $N_s$, which directly influences the computational complexity of $f_A$, $f_S$, and $f_D$.
While we prioritized a comprehensive scene input to maximize predictive performance in this work, our approach still achieves a 75\% reduction in scene input size compared to the complete map.
Furthermore, incorporating additional domain-specific constraints--such as flight-state information (\eg whether an aircraft has already landed or is scheduled for departure)--presents a direct opportunity for input token reduction and latency optimization.

\section{CONCLUSIONS}
We present \mn, a trajectory prediction architecture that adapts advancements in autonomous driving to the unique requirements of airport surface operations.
The core of our approach is a novel PRS-based scene context sampling mechanism, which enables the model to effectively process heterogeneous airfield map context across the varying spatial scales of movements.
Extensive evaluations on the Amelia-10 benchmark demonstrate a significant performance improvement over the existing baseline, particularly in safety-critical scenarios.

Our approach provides a robust technical foundation for aviation-specific motion forecasting, highlighting that deep learning methods tailored to the specific problem domain can overcome the limitations of standard architectures.
Future research, conducted in collaboration with air traffic control authorities, will focus on evolving these predictive capabilities into integrated, automated safety systems designed to enhance the operational safety and efficiency of airport surface movements.

\bibliography{chapters/_ref}

\end{document}